\documentclass[11pt]{article}
\usepackage{amssymb}
\usepackage{amsfonts}
\usepackage{amsmath}
\usepackage{latexsym}
\usepackage{epsfig}

\newcommand{\poly}{\mathrm{poly}}

\newcommand{\eps}{{\epsilon}}

\newcommand{\ignore}[1]{}

\newcommand{\strutje}{\rule[-.25cm]{0cm}{.7cm}}

\newcommand{\PP}{\mathsf{PP}}
\newcommand{\PNP}{\mathsf{P^{NP}}}

\newtheorem{theorem}{Theorem} 
\newtheorem{fact}[theorem]{Fact}
\newtheorem{observation}[theorem]{Observation}

\newtheorem{claim}[theorem]{Claim}
\newtheorem{definition}[theorem]{Definition}
\newtheorem{corollary}[theorem]{Corollary}

\newenvironment{proof}{\noindent \textbf{Proof:}}{\hfill{$\Box$}}

\title{Toward Attribute Efficient Learning Algorithms}
\ignore{
OR \\
       Learning Decision Lists of Length $k$ using
       $2^{\tilde{O}(k^{1/3})}$ Examples OR \\ 
       On Learning Decision Lists Attribute Efficiently OR \\ 
       Learning Decision Lists Attribute Efficiently via Polynomial Threshold Functions OR \\
       Learning Decision Lists using $2^{\tilde{O}(k^{1/3})}$ Samples OR \\
       Learning Decision Lists via Polynomial Threshold Functions OR \\
       A Subexponential Algorithm for Learning Decision Lists Attribute Efficiently OR \\
       some other lame title
}

\author{Adam R. Klivans\thanks{Supported by an NSF Mathematical
Sciences Postdoctoral Research Fellowship.}\\
Divsion of Engineering and Applied Sciences\\ 
Harvard University\\ Cambridge, MA 02138 \\{\tt klivans@eecs.harvard.edu}
\and Rocco A.\ Servedio\\ 
Department of Computer Science\\
Columbia University\\ 
New York, NY 10027\\ {\tt rocco@cs.columbia.edu} }

\date{}

\begin{document}

\setcounter{page}{0}

\maketitle

\begin{abstract}

We make progress on two important problems regarding attribute
efficient learnability.  

First, we give an algorithm for learning decision
lists of length $k$ over $n$ variables using $2^{\tilde{O}(k^{1/3})}
\log n$ examples and time $n^{\tilde{O}(k^{1/3})}$. This is the first
algorithm for learning decision lists that has both subexponential
sample complexity and subexponential running time in the relevant
parameters.  Our approach establishes a relationship between attribute
efficient learning and polynomial threshold functions and is based on
a new construction of low degree, low weight polynomial threshold
functions for decision lists. For a wide range of parameters our
construction matches a 1994 lower bound due to Beigel for the
ODDMAXBIT predicate and gives an essentially optimal tradeoff between
polynomial threshold function degree and weight.  

Second, we give an
algorithm for learning an unknown parity function on $k$ out of $n$
variables using $O(n^{1-1/k})$ examples in time polynomial in $n$. For
$k=o(\log n)$ this yields a polynomial time algorithm with
sample complexity $o(n)$.  This is the first polynomial time algorithm
for learning parity on a superconstant number of variables with
sublinear sample complexity.

\end{abstract}

\thispagestyle{empty}

\newpage

\section{Introduction}

\subsection{Attribute Efficient Learning}

A central goal in machine learning is to design efficient, effective
algorithms for learning from small amounts of data.  An obstacle to
achieving this goal is that learning problems are often characterized by
an abundance of {\em irrelevant information}.  In many learning problems
each data point is naturally viewed as a high dimensional vector of
attribute values;  as a motivating example, in a natural language domain a
data point representing a text document may be a vector of word
frequencies over a lexicon of 100,000 words (attributes).  A newly
encountered word in a corpus may typically have a simple definition which
uses only a dozen or so words from the entire lexicon.  One would like to
be able to learn the meaning of such a word using a number of examples
which is closer to a dozen (the actual number of relevant attributes) than
to 100,000 (the total number of attributes).

Towards this end, an important goal in machine learning theory is to
design {\em attribute efficient} algorithms for learning various classes
of Boolean functions.  A class ${\cal C}$ of Boolean functions over $n$
variables $x_1,\dots,x_n$ is said to be {\em attribute-efficiently
learnable} if there is a poly$(n)$ time algorithm which can learn any
function $f \in C$ using a number of examples which is polynomial in the
``size'' (description length) of the function $f$ to be learned, rather
than in $n$ (the number of features in the domain over which learning
takes place).  (Note that the running time of the learning algorithm must
in general be at least $n$ since each example is an $n$-bit vector.)  
Thus an attribute efficient learning algorithm for, say, the class of
Boolean conjunctions must be able to learn any Boolean conjunction of $k$
literals over $x_1,\dots,x_n$ using poly$(k,\log n)$ examples, since $k
\log n$ bits are required to specify such a conjunction.

\subsection{Decision Lists}

A longstanding open problem in machine learning, posed first by Blum in
1990 \cite{Blum:90,Blum:96,BHL:95,BlumLangley:97} and again by 
Valiant in 1998
\cite{Valiant:99}, is to determine whether or not there exist attribute
efficient algorithms for learning {\em decision lists}.  A decision list
is essentially a nested ``if-then-else'' statement (we give a precise
definition in Section \ref{sec:prelims}).

Attribute efficient learning of decision lists is of both theoretical and
practical interest. Blum's motivation for considering the problem came
from the {\em infinite attribute model} \cite{Blum:90}; in this model
there are infinitely many attributes but the concept to be learned depends
on only a small number of them, and each example consists of a finite list
of active attributes.  Blum {\em et al}. \cite{BHL:95} showed that for a
wide range of concept classes (including decision lists)  attribute
efficient learnability in the standard $n$-attribute model is equivalent
to learnability in the infinite attribute model.  Since simple classes
such as disjunctions and conjunctions are attribute efficiently learnable
(and hence learnable in the infinite attribute model), this motivated Blum
\cite{Blum:90} to ask whether the richer class of decision lists is thus
learnable as well.\footnote{ Additional motivation comes from the fact
that decision lists have such a simple algorithm in the PAC model.}
Several researchers have subsequently considered this problem, see e.g.
\cite{Blum:96,BlumLangley:97,DhagatHellerstein:94, NevoElYaniv:02,
Servedio:99stoc}; we summarize some of this previous work in Section
\ref{sec:prevdl}.
    
From an applied perspective, Valiant \cite{Valiant:99} relates the
problem of learning decision lists attribute efficiently to the question
``how can human beings learn from small amounts of data in the presence of
irrelevant information?'' He points out that since decision lists play an
important role in various models of cognition, a first step in
understanding this phenomenon would be to identify efficient algorithms
which learn decision lists from few examples. Due to the lack of progress
in developing such algorithms for decision lists, Valiant suggests that
models of cognition should perhaps focus on ``flatter" classes of
functions such as projective DNF \cite{Valiant:99}.

\subsection{Parity Functions}

Another outstanding challenge in machine learning is to determine whether 
there exist attribute efficient algorithms for learning {\em parity
functions}.  The parity function
on a set of 0/1-valued variables $x_{i_1},\ldots,x_{i_k}$ is equal to $x_{i_1} + \cdots
+ x_{i_k}$ modulo 2.  As with the class of decision lists, a simple PAC learning
algorithm is known for the class of parity functions but no attribute efficient 
PAC learning algorithm is known.
Learning parity
functions plays an important rule in Fourier learning methods
\cite{MOS:03} and is closely related to  decoding random linear codes \cite{BKW:00}.
Both A. Blum \cite{Blum:96} and Y. Mansour \cite{Man:02} cite
attribute efficient learning of parity functions as an important open
problem.

\ignore{
Given a set of examples labelled according to an unknown parity
function on $k$ out of $n$ variables, we wish to find an approximation
to the unknown parity in polynomial time using as few examples as
possible.  The well known solution to this problem views these
examples as a set of linear equations mod $2$ in $n$ variables and
solves the set of equations to come up with a consistent
hypothesis. Note, however, that we must take $\Omega(n)$ examples to
achieve a solution which has good generalization error, as a solution
to a system of $m$ equations over $n$ variables may contain
$\min(m,n)$ non-zero entries.  An attribute efficient algorithm for
learning parity should require a number of examples polynomially
related to $k$ and $\log n$ (information theoretically we should only
need $O(k \log n)$ examples).
}

\subsection{Our Results: Decision Lists}

We give the first learning algorithm for decision lists that is
subexponential in both sample complexity (in the relevant parameters $k$
and $\log n$) and running time (in the relevant parameter $k$).  Our
results demonstrate for the first time that it is possible to
simultaneously avoid the ``worst case'' in both sample complexity and
running time, and thus suggest that it may indeed be possible to learn
decision lists attribute efficiently. \ignore{We consider this to be the
first evidence that decision lists can be learned attribute efficiently.
\\}

Our main learning result for decision lists is:  

\begin{theorem} \label{thm:main} There is an algorithm for learning
decision lists over $\{0,1\}^n$ which, when learning a decision list
of length $k$, has mistake bound\footnote{Throughout this
section we use ``sample complexity'' and ``mistake bound''
interchangeably; as described in Section \ref{sec:prelims}
these notions are essentially identical.}
$2^{\tilde{O}(k^{1/3})}\log n$ and runs  in time
$n^{\tilde{O}(k^{1/3})}$.
\end{theorem}

We prove Theorem \ref{thm:main} in two parts; first we generalize
Littlestone's well known Winnow algorithm \cite{Littlestone:88}
for learning 
linear threshold functions to learn {\em polynomial
threshold functions.} In previous learning results, polynomial threshold
functions are learned by applying techniques from linear programming: a
Boolean function computed by a polynomial threshold function of degree $d$ can
be learned in time $n^{O(d)}$ by using polynomial time linear programming
algorithms such as the Ellipsoid algorithm 
(see e.g. \cite{KlivansServedio:01}).
\ignore{via a linear programming solver, such as the
Ellipsoid algorithm.}
In contrast, we use the Winnow algorithm to learn polynomial threshold functions.
Winnow learns using few examples in a small amount of time
provided that the degree of the polynomial
is low and the integer coefficients of the polynomial are not too large:
\ignore{As opposed to general
linear programming solvers, Winnow can learn in an attribute efficient
manner:}

\begin{theorem} \label{thm:win}
Let ${\cal C}$ be a class of Boolean functions over
$\{0,1\}^n$ with the property that each $f \in {\cal C}$ has a polynomial
threshold function of degree at most $d$ and weight at most $W.$ Then
there is an online learning algorithm for ${\cal C}$ which runs in $n^d$
time per example and has mistake bound $O(W^{2} \cdot d \cdot \log n).$
\end{theorem}

At this point we have reduced the problem of learning decision lists
attribute efficiently to the problem of representing decision lists with
polynomial threshold functions of low weight and low degree. To this end
we prove

\begin{theorem} \label{thm:ptf} Let $L$ be a decision list of length $k$.
Then $L$ is computed by a polynomial threshold function of degree
$\tilde{O}(k^{1/3})$ and weight $2^{\tilde{O}(k^{1/3})}$.  \end{theorem}
Theorem \ref{thm:main} follows directly from Theorems \ref{thm:win}
and \ref{thm:ptf}.

Polynomial threshold function constructions have recently been used
to obtain the fastest known algorithms for a range
of important learning problems such as learning DNF formulas
\cite{KlivansServedio:01}, intersections of halfspaces \cite{KOS:02}, 
and Boolean formulas of superconstant depth \cite{OdonnellServedio:03a}.  
For each of these learning problems the sole goal was to obtain
fast learning algorithms, and hence the only parameter of interest in 
these polynomial threshold function constructions is their degree, 
since degree bounds translate directly into running time bounds for
learning algorithms (see e.g. \cite{KlivansServedio:01}).
In contrast, for the decision list problem we are interested in 
both the running time and the number of examples required for learning.
Thus we must bound both the degree and the {\em weight} 
(magnitude of integer coefficients) of the polynomial threshold 
functions which we use.

Our polynomial threshold function construction is essentially optimal in
the tradeoff between degree and weight which it achieves.  In 1994 Beigel
gave a lower bound showing that any degree $d$ polynomial threshold
function for a particular decision list must have weight
$2^{\Omega(n/d^{2})}$. For $d = n^{1/3}$, Beigel's lower bound implies
that the construction stated in Theorem \ref{thm:ptf} is essentially
optimal.  Furthermore, for any decision list $L$ of length $n$ and any
$d \leq n^{1/3}$, we will in fact construct polynomial threshold functions 
of degree $d$ and weight $2^{\tilde{O}(n/d^{2})}$ computing $L$. 
Beigel's lower bound thus implies that our degree $d$ polynomial threshold 
functions are of roughly optimal weight
for all $d \leq n^{1/3},$ and hence strongly suggests that our 
analysis is the best possible for the algorithm we use.

\subsection{Our Results: Parity Functions}

For parity functions, we give an $O(n^3)$ time algorithm which can 
learn an unknown parity on $k$ variables out of $n$ using $O(n^{1-1/k})$ examples.
For values of $k = o(\log n)$ the sample complexity of
this algorithm is $o(n)$. This is the first algorithm for learning
parity on a superconstant number of variables with sublinear sample
complexity.

The standard PAC learning algorithm for learning an unknown parity function
is based on viewing a set of $m$ labelled examples as a system of $m$ linear equations modulo 2.
Using Gaussian elimination it is possible to solve the system and find 
a consistent parity function.  It can be shown that the solution thus
obtained is a ``good'' hypothesis if its weight (number of nonzero entries)
is small relative to $m$, the number of examples.  However, using Gaussian elimination
can result in a solution of weight as large as  
$\min(m,n)$ even if $k$ (the number of variables in the target parity) is very small.
Thus in order for this approach to give a successful learning algorithm, it is necessary to 
use $m = \Omega(n)$ examples regardless of the value of $k$. 
In contrast, observe that an attribute efficient algorithm for
learning a parity of length $k$ should use only poly$(k,\log n)$ examples.

Our algorithm works by finding a ``low weight'' solution to a system of
$m$ linear equations.  We prove that with high probability we can find a solution of weight
$O(n^{1-1/k})$ irrespective of $m$.  Thus by taking $m$ to be only slightly larger
than $n^{1 - 1/k}$ we have that our solution is a ``good'' hypothesis.

\subsection{Previous Results: Decision Lists} \label{sec:prevdl}

In previous work several algorithms with different performance bounds (in
terms of running time and number of examples used) have been given for
learning decision lists.

\begin{itemize}

\item Rivest \cite{Rivest:87} gave the first algorithm for learning
decision lists in Valiant's PAC model of learning from random examples.  
Littlestone \cite{Blum:96} subsequently gave an analogue of Rivest's
algorithm in the online learning model. The algorithm can learn any
decision list of length $k$ in $O(kn^2)$ time using $O(kn)$ examples.

\item A brute-force approach to learning decision lists of length $k$ is
to maintain a collection of all such lists which are consistent with the
examples seen so far, and to predict at each stage using majority vote
over the surviving hypotheses. This ``halving algorithm'' (proposed in
various forms by Barzdin and Freivald \cite{BarzdinFreivald:72}, Mitchell
\cite{Mitchell:82}, and Angluin \cite{Angluin:88}) can learn decision
lists of length $k$ using only $O(k \log n)$ examples, but the running
time is $n^{O(k)}.$

\item Several researchers \cite{Blum:96,Valiant:99} have observed that
Littlestone's well-known Winnow algorithm \cite{Littlestone:88} can learn
decision lists of length $k$ from $2^{O(k)} \log n$ examples in time
$2^{O(k)} n \log n$. This follows from the observation that decision lists
of length $k$ can be viewed as linear threshold functions with integer
coefficients of magnitude $2^{\Theta(k)}$. We note that our algorithm in
this paper always has improved sample complexity over the basic Winnow
algorithm, and for $k \geq (\log n)^{3/2}$ our approach improves on the
time complexity of Winnow as well.

\item Finally, several researchers have considered the special
case of learning a decision list of length $k$ over $n$ variables
in which the output bits of the decision list have at most $D$
alternations. Valiant \cite{Valiant:99}
and Nevo and El-Yaniv \cite{NevoElYaniv:02}
have given refined analyses of Winnow's performance for this
special case, and Dhagat and Hellerstein \cite{DhagatHellerstein:94} 
have also studied this problem.  However, for the general case
in which $D$ can be as large as $k,$ the results thus obtained
do not improve on the straightforward Winnow analysis 
described in the previous bullet.

\end{itemize}
These previous algorithmic results are summarized in Figure 1.  We observe
that all of these earlier algorithms have an exponential dependence on the
relevant parameter(s) ($k$ and $\log n$ for sample complexity, $k$ for
running time)  for either the running time or the sample complexity.

\begin{table}[h]
\centerline{
\begin{tabular}{|l|l|l|} \hline
\strutje Reference: & Number of examples: & Running time: \\
\hline\hline
\strutje Rivest / Littlestone
& $ O(kn)$ 
& $ O(kn^2)  $ \\ \hline
\strutje Halving algorithm
& $ O(k \log n)$
& $ n^{O(k)} $ \\ \hline
\strutje Winnow algorithm
& $2^{O(k)} \log n$ 
& $2^{O(k)}n \log n$  \\ \hline
\strutje This Paper
& $ 2^{\tilde{O}(k^{1/3})}\log n $
& $ n^{\tilde{O}(k^{1/3})} $  \\ \hline
\end{tabular}
}
\caption{Comparison of known algorithms for 
learning decision lists of length $k$ on $n$ variables.
}
\label{table:results} 
\end{table}

\subsection{Previous Results: Parity Functions}

Little previous work has been published on learning parity
functions attribute efficiently in the PAC model.  The standard PAC learning
algorithm for parity (based on solving a system of linear equations) is due
to Helmbold {\em et al.\@} \cite{HSW:92}; however as described above this
algorithm is not attribute efficient since it uses $\Omega(n)$ examples.

Several authors have considered learning parity attribute efficiently in a model 
where the learner is allowed to make membership queries.  Attribute efficient
learning is easier in this framework since membership queries can help identify relevant variables.
Blum et al. \cite{BHL:95} give a randomized polynomial time membership-query
algorithm for learning parity on $k$ variables using only $O(k \log
n)$ examples.  These results were later
refined by Uehara {\em et al.} \cite{UTW:97}.

\subsection{Organization}

In Section \ref{sec:prelims} we give the necessary background on
online learning and polynomial threshold functions. In Section
\ref{sec:winnow} we show how known results from learning theory enable
us to reduce the decision list learning problem to a problem of
finding suitable polynomial threshold function representations of
decision lists. In Sections \ref{subsec:outer} and \ref{subsec:inner}
we give two different proofs of a weak tradeoff between degree and
weight for polynomial threshold function representations of decision
lists, and in Section \ref{subsec:compose} we combine these techniques
to prove Theorem \ref{thm:ptf}. In Section \ref{sec:decisiontree} we
show how to apply our techniques to give a tradeoff between sample
complexity and running time for learning decision trees. In Section
\ref{sec:discuss} we discuss the connection with Beigel's ODDMAXBIT
lower bound and related issues.  In Section \ref{sec:parity} we give
our new algorithm for learning parity functions, and in Section
\ref{sec:future} we suggest directions for future work.

\section{Preliminaries} \label{sec:prelims}

Attribute efficient learning has been chiefly studied in the {\em on-line
mistake-bound} model of concept learning which was introduced in
\cite{Littlestone:88,Littlestone:89}.  In this model learning proceeds in
a series of trials, where in each trial the learner is given an unlabelled
boolean example $x \in \{0,1\}^n$ and must predict the value $f(x)$ of the
unknown target function $f.$ After each prediction the learner is given
the true value of $f(x)$ and can update its hypothesis before the next
trial begins.  The {\em mistake bound} of a learning algorithm on a target
concept $c$ is measured by the worst-case number of mistakes that the
algorithm makes over all (possibly infinite) sequences of examples, and
the mistake bound of a learning algorithm on a concept class (class of
Boolean functions) $C$ is the worst-case mistake bound across all
functions $f \in C.$ The running time of a learning algorithm $A$ for a
concept class $C$ is defined as the product of the mistake bound of $A$ on
$C$ times the maximum running time required by $A$ to evaluate its
hypothesis and update its hypothesis in any trial.

Our main interests in this paper are the classes of {\em decision
lists} and {\em parity functions}.

A decision list $L$ of length $k$ over the Boolean variables
$x_1,\dots,x_n$ is represented by a list of $k$ pairs and a bit
$$
(\ell_1,b_1),(\ell_2,b_2),\dots,(\ell_k,b_k),b_{k+1}
$$
where each $\ell_i$ is a literal and each $b_i$ is either $-1$ or $1.$
Given any $x \in \{0,1\}^n,$ the value of $L(x)$ is $b_i$ if $i$ is the
smallest index such that $\ell_i$ is made true by $x$; if no $\ell_i$ is
true then $L(x)=b_{k+1}.$

A parity function of length $k$ is defined by a set of variables $S
\subset \{x_{1},\ldots,x_{n}\}$ such that $|S| = k$. The 
parity function $\chi_{S}(x)$ takes value $1$ on inputs which set
an even number of variables in $S$ to $1$ and takes value $-1$ on
inputs which set an odd number of variables in $S$ to $1.$

Given a concept class $C$ over $\{0,1\}^n$ and a Boolean function $f \in
C,$ let size$(f)$ denote the description length of $f$ under some
reasonable encoding scheme.  (Note that if $f$ has $r$ relevant variables
then size$(f)$ will be at least $r \log n$ since this many bits are
required just to specify which variables are relevant).  We say that a
learning algorithm $A$ for $C$ in the mistake-bound model is {\em
attribute-efficient} if the mistake bound of $A$ on any concept $c \in C$
is polynomial in size$(f).$ In particular, the description length of a
length $k$ decision list (parity) is $O(k \log n)$, and thus we would ideally like
to have an algorithm which learns decision lists (parities) of length $k$ with a
mistake bound of poly$(k,\log n)$ and runs in time poly$(n).$

(We note here that attribute efficiency has also been studied in other
learning models, namely Valiant's Probably Approximately Correct (PAC)
model of learning from random examples.  Standard conversion techniques
are known \cite{Angluin:88,Haussler:88b,Littlestone:89b}
which can be used to
transform any mistake bound algorithm into a PAC learning algorithm.  
This transformation essentially preserves the running time of the mistake
bound algorithm, and the sample size required by the PAC algorithm is
essentially the mistake bound. Thus, positive results for mistake bound
learning, such as those we give for decision lists in this paper, directly yield
corresponding positive results for the PAC model.)

Finally, our results for decision lists are achieved by a careful
analysis of {\em polynomial threshold functions}.  Let $f$ be a
Boolean function $f:\{0,1\}^{n} \to \{-1,1\}$ and let $p$ be a
polynomial in $n$ variables with integer coefficients. Let $d$ denote
the degree of $p$ and let $W$ denote the sum of the absolute values of
$p$'s integer coefficients. If the sign of $p(x)$ equals $f(x)$ for
every $x \in \{0,1\}^n,$ then we say that $p$ is a {\em polynomial
threshold function} of degree $d$ and weight $W$ for $f.$

\section{Expanded-Winnow: Learning Polynomial Threshold Functions} \label{sec:winnow}

Littlestone introduced the online Winnow algorithm in 1988 and showed
that it can attribute efficiently learn Boolean conjunctions,
disjunctions, and low weight linear threshold functions.  Throughout
its execution Winnow maintains a linear threshold function as its
hypothesis; at the heart of the algorithm is a novel update rule which
makes a {\em multiplicative} update to each coefficient of the
hypothesis (rather than an additive update as in the Perceptron
algorithm) each time a mistake is made.  Since its introduction Winnow
has been intensively studied from both applied and theoretical
standpoints (see
e.g. \cite{Blum:97,GoldingRoth:99,KWA:97,Servedio:02sicomp}) and
multiplicative updates have become widespread in machine learning
algorithms.

The following theorem (which, as noted in \cite{Valiant:99}, is implicit
in Littlestone's analysis in \cite{Littlestone:88}) gives a 
mistake bound for Winnow when learning linear threshold functions:

\begin{theorem} \label{thm:winbound}
Let $f(x)$ be the linear threshold function 
sign$(\sum_{i=1}^{n} w_{i}x_{i} - \theta)$ 
where $\theta$ and $w_{1},\ldots,w_{n}$ are
integers. Let $W = \sum_{i=1}^{n} |w_{i}|$. Then 
Winnow learns $f(x)$ with mistake bound $O(W^{2} \log n)$,
and uses $n$ time steps per example.
\end{theorem}

We will use a generalization of the Winnow algorithm, called
Expanded-Winnow, to learn {\em polynomial} threshold functions of
degree at most $d.$ Our generalization introduces $\sum_{i=1}^{d} {n
\choose d}$ new variables (one for each monomial of degree up to $d$)
and runs Winnow to learn a linear threshold function over these new
variables.  More precisely, in each trial we convert the $n$-bit
received example $x=(x_1,\dots,x_n)$ into a $\sum_{i=1}^d {n \choose
d}$ bit expanded example (where the bits in the expanded example
correspond to monomials over $x_1,\dots,x_n$), and we give the
expanded example to Winnow.  Thus the hypothesis which Winnow
maintains -- a linear threshold function over the space of expanded
features -- is a polynomial threshold function of degree $d$ over the
original $n$ variables $x_1,\dots,x_n.$ Theorem \ref{thm:win}, which
follows directly from Theorem \ref{thm:winbound}, summarizes the
performance of Expanded-Winnow:

\medskip

\noindent {\bf Theorem \ref{thm:win}}
{\em Let ${\cal C}$ be a class of Boolean functions over
$\{0,1\}^n$ with the property that each $f \in {\cal C}$ has a polynomial
threshold function of degree at most $d$ and weight at most $W.$ Then
Expanded-Winnow algorithm runs in $n^d$
time per example and has mistake bound $O(W^{2} \cdot d \cdot \log n)$ for
${\cal C}.$
} \\

Theorem \ref{thm:win} shows that the degree of a polynomial threshold
function corresponds to Expanded-Winnow's running time, and the weight of
a polynomial threshold function corresponds to its sample complexity.

\section{Constructing Polynomial Threshold Functions for Decision Lists}

In previous constructions of polynomial threshold functions for
computational learning theory applications
\cite{KlivansServedio:01,KOS:02,OdonnellServedio:03a} the sole goal has
been to minimize the {degree} of the polynomials regardless of the size of
the coefficients.  As an extreme example, the construction of
\cite{KlivansServedio:01} of $\tilde{O}(n^{1/3})$ degree polynomial
threshold functions for DNF formulae yields polynomials whose coefficients
can be {\em doubly exponential} in the degree. In contrast, 
given Theorem \ref{thm:win} we must now
construct polynomial threshold functions that have low degree and low
weight.

We give two constructions of polynomial threshold functions for decision lists, each of which
has relatively low degree \ignore{($k^{1/2}$)} 
and relatively low weight. 
\ignore{($2^{\tilde{O}(k^{1/2})}$).}  
We then combine 
these approaches to achieve an optimal construction with improved bounds on both
degree and weight.\ignore{with degree $k^{1/3}$
and weight $2^{\tilde{O}(k^{1/3})}.$}

\subsection{Outer Construction} \label{subsec:outer}

Let $L$ be a decision list of length $k$ over variables $x_1,\dots,x_k.$
We first give a simple construction of a degree $h$, weight ${\frac {2k}
h}2^{(k/h + h)}$ polynomial threshold function for $L$ which is based on
breaking the list $L$ into sublists.  We call this construction the
``outer construction" since we will ultimately combine this construction
with a different construction for the ``inner'' sublists.

We begin by showing that $L$ can be expressed as a threshold of {\em
modified decision lists} which we now define.  The set ${\cal B}_h$ of
modified decision lists is defined as follows:
each function in ${\cal B}_h$ is a decision list
$(\ell_1,b_1),(\ell_2,b_2),\dots, (\ell_h,b_h),0$ where each $\ell_i$ is
some literal over $x_1,\dots,x_n$ and each $b_i \in \{-1,1\}.$ Thus the
only difference between a modified decision list $f \in {\cal B}_h$ and a
normal decision list of length $h$ is that the final output value is
$0$ rather than $b_{h+1} \in \{-1,+1\}.$

Without loss of generality we may suppose that the list $L$ is
$(x_1,b_1),\dots,(x_k,b_k),b_{k+1}.$ We break $L$ sequentially into $k/h$
blocks each of length $h$. Let $f_{i} \in {\cal B}_h$ be the modified
decision list which corresponds to the $i$-th block of $L,$ i.e. $f_i$ is
the list $(x_{(i-1) h + 1},b_{(i-1)h+1}),\ldots, (x_{(i+1)
h},b_{(i+1)h}),0$.  Intuitively $f_{i}$ computes the $i$th block of $L$
and equals $0$ only if we ``fall of the edge" of the $i$th block. We then
have the following straightforward claim:

\begin{claim} \label{cla:outer}
The decision list $L$ is eqivalent to 
\begin{eqnarray}
\mbox{sign}\left(\sum_{i=1}^{k/h}
2^{k/h - i + 1} f_{i}(x) \ + \  b_{k+1} \right). \label{eq:outer}
\end{eqnarray}
\end{claim}
\begin{proof}
Given an input $x \neq 0^k$ let $r=(i-1)h + c$ be the first index such that $x_r$ is satisfied.
It is easy to see that $f_j(x) = 0$ for $j<i$ and hence the value in 
(\ref{eq:outer}) is $2^{k/h - i + 1}b_{r} + \sum_{j=i+1}^{k/h}
2^{k/h - j + 1} f_{j}(x) \ + \  b_{k+1}$, 
the sign of which is easily seen to be $b_r.$
Finally if $x=0^k$ then the argument to (\ref{eq:outer}) is $b_{k+1}$.
\end{proof}

\medskip \noindent {\bf Note:}  It is easily seen that we can replace
the $2$ in formula (\ref{eq:outer}) by a 3; this will prove
useful later.

\medskip

As an aside, note that Claim \ref{cla:outer} can already be used to obtain a tradeoff
between running time and sample complexity for learning decision lists.
The class ${\cal B}_h$ contains at most $(4n)^h$ functions.  
Thus as in Section \ref{sec:winnow}
it is possible to run the Winnow algorithm using the functions in ${\cal B}_h$ as the base features
for Winnow.  (So for each example $x$ which it receives, the algorithm would first compute
the value of $f(x)$ for each $f \in {\cal B}_h$, and would then use this vector of $(f(x))_{f \in {\cal B}_h}$
values as the example point for Winnow.)  A direct analogue of Theorem 
\ref{thm:win} now implies
that Expanded-Winnow (run over this expanded feature space of functions from 
${\cal B}_h$) can be used to learn 
$L_k$ in time $n^{O(h)}2^{O(k/h)}$ with mistake bound $2^{O(k/h)} h \log n$.

However, it will be more useful for us to obtain a polynomial threshold function for $L$.  We
can do this from Claim \ref{cla:outer} as follows:

\begin{theorem} \label{thm:outer}
Let $L$ be a decision list of length $k$.  Then for any $h < k$
we have that $L$ is computed by a
polynomial threshold function of degree $h$ 
and weight $4 \cdot 2^{k/h + h}$.
\end{theorem}

\begin{proof}
Consider the first modified decision list $f_1 = (\ell_1,b_1),(\ell_2,b_2),\dots,(\ell_h,b_h),0$ 
in the expression (\ref{eq:outer}).  For $\ell$ a literal let $\tilde{\ell}$ denote $x$
if $\ell$ is an unnegated variable $x$ and let $\tilde{\ell}$ denote $1-x$ if 
if $\ell$ is a negated variable $\overline{x}.$ 
We have that for all $x \in \{0,1\}^h$, $f_1(x)$ is computed exactly by
the polynomial
$$
f_1(x) = \tilde{\ell}_1b_1 + (1-\tilde{\ell}_1)\tilde{\ell}_2 b_2 + 
(1-\tilde{\ell}_1)(1-\tilde{\ell}_2)\tilde{\ell}_3 b_3 + \cdots + 
(1-\tilde{\ell}_1)\cdots(1-\tilde{\ell}_{h-1})\tilde{\ell}_h b_h.
$$
This polynomial has degree $h$ and has weight at most $2^{h+1}.$
Summing these polynomial representations for $f_1,\dots,f_{k/h}$ 
as in (\ref{eq:outer}) we see
that the resulting polynomial threshold function given by (\ref{eq:outer})
has degree $h$ and weight at most $2^{k/h + 1} \cdot 2^{h+1} = 
4 \cdot 2^{k/h + h}.$
\end{proof}

\medskip

Specializing to the case $h=\sqrt{k}$ we obtain:

\begin{corollary} \label{cor:outer}
Let $L$ be a decision list of length $k$.
Then $L$ is computed by a polynomial threshold function of
degree $k^{1/2}$ and weight $4 \cdot 2^{2k^{1/2}}.$
\end{corollary}

We close this section by observing that an intermediate result 
of \cite{KlivansServedio:01} can be used to give an alternate proof 
of Corollary \ref{cor:outer} with slightly weaker parameters; 
see Appendix \ref{ap:alt}.

\subsection{Inner Approximator} \label{subsec:inner}

In this section we construct low degree, low weight 
polynomials which approximate (in the $L_\infty$ norm)
the modified decision lists from the previous subsection.  Moreover,
the polynomials we construct 
are exactly correct on inputs which ``fall off the end'':
\ignore{
We refer to these modified decision lists as the ``inner'' decision lists.
The construction is stronger than a polynomial threshold function; 
the polynomial we give for an inner decision list is actually
a good approximator with respect to the
$L_{\infty}$ norm (and is exactly right on the input $0^h$):
}

\begin{theorem} \label{thm:inner}
Let $f \in {\cal B}_h$ be a modified decision list of length $h$ 
(without loss of generality we may assume that $f$ is
$(x_1,b_1),\dots,(x_h,b_h),0$).
Then there is a degree $2\sqrt{h}\log{h}$
polynomial $p$ such that 
\begin{itemize}
\item for every input $x \in \{0,1\}^h$ we have $|p(x) - f(x)| \leq 1/h$. 
\item $p(0^h) = f(0^h) = 0$.
\end{itemize}
\end{theorem}
\begin{proof}
As in the proof of Theorem \ref{thm:outer} we have that
\[ f(x) = b_{1}x_{1} + b_{2}(1-x_{1})x_{2} + \cdots + 
b_{h}(1-x_{1})\cdots(1-x_{h-1})x_{h}. 
\]
We will construct a lower (roughly $\sqrt{h}$) degree polynomial which 
closely approximates $f$.  Let $T_{i}$ denote $(1-x_1)\dots(1-x_{i-1})x_i$,
so we can rewrite $f$ as
\[ f(x) = b_{1}T_{1} + b_{2}T_{2} + \cdots + b_{h}T_{h}. \]

We approximate each $T_i$ separately as follows:
set $A_{i}(x) = h-i  + x_{i} + \sum_{j=1}^{i-1} (1 - x_{j})$.
Note that for $x \in \{0,1\}^h,$ we have
$T_i(x) = 1$ iff $A_i(x) = h$ and $T_i(x) = 0$
iff $0 \leq A_i(x) \leq h-1.$
Now define the polynomial 
$$
Q_{i}(x) = q \left(A_{i}(x)/h \right)  \mbox{~~~~~where~~~~~}
q(y) = C_d\left(y \left(1 + 1/h \right) \right).
$$

\noindent As in \cite{KlivansServedio:01},
here $C_{d}(x)$ is the $d$th Chebyshev polynomial of the
first kind (a univariate polynomial of degree $d$) 
with $d$ set to $\lceil \sqrt{h} \rceil$. 
We will need the following facts about Chebyshev polynomials 
\cite{Cheney:66}: 
\begin{itemize} 
\item $|C_d(x)| \leq 1$ for $|x| \leq 1$ with $C_d(1) = 1;$ 
\item $C_d^\prime(x) \geq d^2$ for $x > 1$ with $C_d^\prime(1) = d^2.$ 
\item The coefficients of $C_{d}$ are integers each of whose
magnitude is at most $2^d$. 
\end{itemize}
These first two facts imply that $q(1) \geq 2$ but $|q(y)| \leq 1$ 
for $y \in [0,1 - {\frac 1 h}].$  We
thus have that $Q_i(x) = q(1) \geq 2$ if $T_i(x) = 1$
and $|Q_i(x)| \leq 1$ if $T_i(x) = 0.$
Now define 
$
P_i(x) = \left({\frac {Q_i(x)}{q(1)}}\right)^{2 \log h}.
$
This polynomial is easily seen to be a good approximator for $T_i$:
if $x \in \{0,1\}^h$ is such that $T_i(x) = 1$ then $P_i(x) = 1$,
and if $x \in \{0,1\}^h$ is such that $T_i(x) = 0$ then
$|P_i(x)| < \left({\frac 1 2}\right)^{2 \log h} < {\frac 1 {h^2}}.$

Now define
$R(x) = \sum_{i=1}^{\ell} b_iP_{i}(x)$ and $p(x) = R(x) - R(0^h).$
\ignore{
We will see that $Q_{i}(x) > 2$ on assignments $x$ for which 
$T_{i}(x)=0$, while $|Q_i(x)|\leq 1$ on assignments for which
$T_{i}(x)$ output $s_{i}$. To
strengthen this separation we define the following polynomial
$P_{i}(x) = (1/\ell^{2}) Q_{i}(x)^{2 \log \ell}$ and to approximate
all of $b$ we set $R(x) = \sum_{i=1}^{\ell} P_{i}(x)$.
}
It is clear that $p(0^h)=0.$ 
We will show that for every input $0^h \neq x \in \{0,1\}^h$ we have 
$|p(x) - f(x)| \leq {1/h}$. Fix some such $x$; let $i$ be the first
index such that $x_i = 1.$  As shown above we have
$P_i(x) = 1.$  Moreover, by inspection of $T_j(x)$ we have that
$T_j(x) = 0$ for all $j \neq i,$  
and hence $|P_j(x)| < {\frac 1 {h^2}}$.  Consequently 
the value of $R(x)$ must lie in $[b_i - {\frac {h-1}{h^2}},
b_i + {\frac {h-1}{h^2}}]$.  Since $f(x) = b_i$ we have that
$p(x)$ is an $L_\infty$ approximator for $f(x)$ as desired.

Finally, it is straightforward to verify that $p(x)$ has the claimed
bound on degree.
\end{proof}

\ignore{
\noindent Now fix any nonzero assignment to the variables $x$ that
causes $b$ to output $1$.  From the definition of $b$ there exists a
unique term $T_{i}$ that is not set to zero by $x$. Then for the
corresponding arithmetization $A_{i}$ we have $A_{i}/i= 1$, so $2 \leq
Q_{i}(x) \leq 2.01 $ and hence $1 \leq P_{i}(x) \leq 1.1$. Similarly
if $x$ causes $b$ to output $-1$ then $-1 \leq P_{i}(x) \leq -.9$. \\

\noindent Let $T_{j}$ be any term that is set to zero by x, and so
$A_{j}(x) \leq 1 - 1/\ell$. Then $|Q_{i}(x)| \leq 1$ and thus
$|P_{i}(x)| \leq 1/\ell^{2}$. Hence for any nonzero assignment $x$,
$|R(x) - b(x)| \leq \mbox{{\bf $\eps$ from cheby approx +
$1/\ell$}}$. Notice also that $|R(\overline{0})| \leq 1/\ell.$ Thus
for any nonzero assignment $x$, $|H(x) - b(x)| \leq 2/\ell$ and
clearly $H(\overline{0}) = 0$. 
}

\medskip

Strictly speaking we cannot discuss the weight of the polynomial
$p$ since its coefficients are rational numbers but not
integers.  However, by multiplying $p$ by a suitable integer
(clearing denominators) we obtain an integer polynomial
with essentially the same properties.
Using the third fact about Chebyshev polynomials from our
proof above, we have that $q(1)$ is a rational number $N_1/N_2$ where
$N_1,N_2$ are each integers of magnitude $h^{O(\sqrt{h})}.$
Each $Q_i(x)$ for $i=1,\dots,h$ can be written as an integer
polynomial (of weight $h^{O(\sqrt{h})}$) divided by $h^{\sqrt{h}}.$
Thus each $P_i(x)$ can be written as 
$\tilde{P}_i(x)/(h^{\sqrt{h}}N_1)^{2 \log h}$ where $\tilde{P}_i(x)$
is an integer polynomial of weight $h^{O(\sqrt{h} \log h)}$.
It follows that $p(x)$ equals $\tilde{p}(x)/C,$ where $C$
is an integer which is at most $2^{O(h^{1/2} \log^2 h)}$
and $\tilde{p}$ is a polynomial with integer coefficients and weight
$2^{O(h^{1/2} \log^2 h)}.$  We thus have

\begin{corollary}
\label{cor:inner}
Let $f \in {\cal B}_h$ be a modified decision list of length $h$.
Then there is an integer polynomial 
$p(x)$ 
of degree $2\sqrt{h}\log{h}$
and weight $2^{O(h^{1/2} \log^2{h})}$ and an integer $C = 
2^{O(h^{1/2} \log^2 h)}$ such that
\begin{itemize}
\item for every input $x \in \{0,1\}^h$ we have $|p(x) - Cf(x)| \leq C/h$.
\item $p(0^h) = f(0^h) = 0$.
\end{itemize}
\end{corollary}

The fact that $p(0^h)$ is exactly 0
will be important in the next subsection when we combine the
inner approximator with the outer construction.

\subsection{Composing the Constructions} \label{subsec:compose}

In this section we combine the two constructions from the previous
subsections to obtain our main polynomial threshold construction:

\begin{theorem} \label{thm:mainptf}
Let $L$ be a decision list of length $k$.  Then for any $h < k$,
$L$ is computed by a polynomial threshold function of degree 
$O(h^{1/2} \log h)$
and weight $2^{O(k/h + h^{1/2}\log^2 h)}.$
\end{theorem}
\begin{proof}
We suppose without loss of generality that $L$ is the decision list
$(x_1,b_1),\dots,(x_k,b_k),b_{k+1}.$
We begin with the outer construction: from the note following
Claim \ref{cla:outer} we have that 
$$L(x) = 
\mbox{sign}\left(C\left[\sum_{i=1}^{k/h}
3^{k/h - i + 1} f_{i}(x) \ + \  b_{k+1} \right]\right)
$$
where $C$ is the value from Corollary \ref{cor:inner} and 
each $f_{i}$ is a modified decision list of length $h$
computing the restriction of $L$ to its $i$th block as defined in
Subsection \ref{subsec:outer}.
Now we use the inner approximator to replace each $Cf_i$ above
by $p_i$, the approximating polynomial from Corollary
\ref{cor:inner}, i.e. consider sign$(H(x))$ where 
$$
H(x) = \sum_{i=1}^{k/h}
(3^{k/h - i + 1} p_{i}(x)) \ + \  Cb_{k+1}.
$$
We will show that sign$(H(x))$
is a polynomial threshold function which computes $L$ correctly
and has the desired degree and weight.

Fix any $x \in \{0,1\}^k.$  If $x=0^k$ then by Corollary
\ref{cor:inner} each $p_i(x)$ is $0$ so $H(x) = C b_{k+1}$ has
the right sign.  
Now suppose that $r=(i-1)h+c$ is the first index such that
$x_r = 1.$  By Corollary \ref{cor:inner}, we have that
\begin{itemize}
\item $3^{k/h - j + 1}p_j(x) = 0$ for $j < i$;
\item $3^{k/h - i + 1}p_i(x)$ differs from $3^{k/h - i + 1}Cb_r$ by at most
$C3^{k/h - i + 1}\cdot {\frac 1 h}$;
\item The magnitude of each value $3^{k/h - j + 1}p_j(x)$ is at most
$C3^{k/h - j + 1}(1 + {\frac 1 h})$ for $j > i.$
\end{itemize}
Combining these bounds,
the value of $H(x)$ differs from $3^{k/h - i + 1}Cb_r$ by at most
$$
C\left(
{\frac {3^{k/h - i + 1}}{h}} + 
\left(1 + {\frac 1 h}\right)
\left[3^{k/h - i} + 3^{k/h - i - 1} + \cdots + 3\right] + 1
\right)
$$
which is easily seen to be less than $C3^{k/h - i + 1}$ in magnitude.
Thus the sign of $H(x)$ equals $b_r$, and consequently sign$(H(x))$ is a
valid polynomial threshold representation for $L(x).$  Finally,
our degree and weight bounds from Corollary \ref{cor:inner}
imply that
the degree of $H(x)$ is $O(h^{1/2} \log h)$ and the weight
of $H(x)$ is $2^{O(k/h) + O(h^{1/2}\log^2 h)}$, and the theorem
is proved.
\end{proof}

\medskip

Taking $h = k^{2/3} / \log^{4/3}k$ in the above theorem we obtain our
main result on representing decision lists as polynomial threshold
functions:

\medskip

\noindent {\bf Theorem \ref{thm:ptf}}
{\em Let $L$ be a decision list of length $k$.  Then 
$L$ is computed by a polynomial threshold function
of degree $k^{1/3} \log^{1/3} k$ and weight
$2^{O(k^{1/3} \log^{4/3} k)}.$
} \\

Theorem \ref{thm:ptf} immediately implies that Expanded-Winnow can learn decision lists of length $k$ using $2^{\tilde{O}(k^{1/3})} \log n$ examples and time $n^{\tilde{O}(k^{1/3})}$.


\section{Application to Learning Decision Trees} \label{sec:decisiontree}

In 1989 Ehrenfeucht and Haussler \cite{EhrenfeuchtHaussler:89} gave an
a time $n^{O(\log s)}$ algorithm for learning decision trees of size
$s$ over $n$ variables. Their algorithm uses $n^{O(\log s)}$ examples,
and they asked if the sample complexity could be reduced to
$\poly(n,s)$.  We can apply our techniques here to give an algorithm
using $2^{\tilde{O}(s^{1/3})} \log n$ examples, if we are willing to
spend $n^{\tilde{O}(s^{1/3})}$ time.

First we need to generalize Theorem \ref{thm:mainptf} for higher order
decision lists. An $r$-decision list is like a standard decision list
but each pair is now of the form $(C_i,b_i)$ where $C_i$ is a
conjunction of at most $r$ literals and as before $b_i = \pm 1$.  The
output of such an $r$-decision list on input $x$ is $b_i$ where $i$ is
the smallest index such that $C_i(x)=1.$

We have the following:
 
\begin{corollary} \label{cor:gdl}
Let $L$ be an $r$-decision list of length $k$. Then for any
$h < k$, $L$ is computed by a polynomial threshold function 
of degree $O(rh^{1/2} \log h)$ and weight 
$2^{r + O(k/h + h^{1/2} \log^2 h)}$.             
\end{corollary}

\begin{proof}
Let $L$ be the $r$-decision list $(C_1,b_1),\dots,(C_k,b_k),b_{k+1}.$
By Theorem \ref{thm:mainptf} there is a polynomial threshold function
of degree $O(h^{1/2} \log h)$ and weight
$2^{O(k/h + h^{1/2} \log^2 h)}$ over the variables $C_1,\dots,C_k.$
Now replace each variable $C_{i}$ by the interpolating polynomial
which computes it exactly as a function from $\{0,1\}^n$ to $\{0,1\}.$
Each such interpolating polynomial has degree $r$ and integer
coefficients of total magnitude at most $2^r$, and the corollary follows.
\end{proof} 

\begin{corollary} \label{cor:learngdl}
There is an algorithm for learning
$r$-decision lists over $\{0,1\}^n$ which, when learning an $r$-decision list
of length $k$, has mistake bound
$2^{\tilde{O}(r + k^{1/3})}\log n$ and runs  in time
$n^{\tilde{O}(rk^{1/3})}$.
\end{corollary}

Now we can apply Corollary \ref{cor:learngdl} to obtain a tradeoff
between running time and sample complexity for learning decision
trees:

\begin{theorem}
Let $D$ be a decision tree of size $s$ over $n$ variables. Then $D$ can be learned using $2^{\tilde{O}(s^{1/3})} \log n$ examples in time $n^{\tilde{O}(s^{1/3})}.$ 
\end{theorem}

\begin{proof}
Blum \cite{Blum:92} has shown that any decision tree of size $s$ is
computed by a $(\log s)$-decision list of length $s.$ Applying
Corollary \ref{cor:learngdl} we thus see that Expanded-Winnow can be
used to learn decision trees of size $s$ over $\{0,1\}^n$ with the
claimed bounds on time and sample complexity.
\end{proof}

\section{Lower Bounds for Decision Lists} \label{sec:discuss}

Here we observe that our construction from
Theorem \ref{thm:mainptf} is essentially optimal in terms of the
tradeoff it achieves between polynomial threshold function degree
and weight.

In \cite{Beigel:94}, Beigel constructs an oracle separating $\PP$ from
$\PNP$. At the heart of his construction is a proof that any low
degree polynomial threshold function for a particular
decision list, called the the $\mathrm{ODDMAXBIT}_{n}$ function,
must have large weights:

\begin{definition}
The $\mathrm{ODDMAXBIT}_{n}$ function on input $x=x_{1},\ldots,x_{n}
\in \{0,1\}^{n}$ equals $(-1)^{i}$ where $i$ is the index of the
first nonzero bit in $x.$
\end{definition}

It is clear that the $\mathrm{ODDMAXBIT}_{n}$ function is 
equivalent to a decision list of length $n$:
$$
(x_1,-1),(x_2,1),(x_3,-1),\dots,(x_n,(-1)^{n}),(-1)^{n+1}.
$$
The main technical theorem which Beigel proves in \cite{Beigel:94}
states that any polynomial threshold function of degree $d$ computing
$\mathrm{ODDMAXBIT}_{n}$ must have weight $2^{\Omega(n/d^{2})}$:

\begin{theorem} \label{thm:beigel}
Let $p$ be a degree $d$ polynomial threshold function with integer
coefficients computing
$\mathrm{ODDMAXBIT}_{n}$. Then  
$w = 2^{\Omega(n/d^{2})}$ where $w$ is the weight of $p.$\footnote{Beigel actually proves something stronger, namely that there must exists a coefficient whose absolute value is at least $2^{\Omega(n/d^{2})}$.}
\end{theorem}
(As stated in \cite{Beigel:94} the bound is actually $w \geq
{\frac 1 s}2^{\Omega(n/d^2)}$ where $s$ is the number of nonzero
coefficients in $p$.  Since $s \leq w$ this implies the result
as stated above.)

A lower bound of $2^{\Omega(n)}$ 
on the weight of any linear threshold function ($d=1$) for
$\mathrm{ODDMAXBIT}_n$ has long been known \cite{MyhillKautz:61};
Beigel's proof generalizes this
lower bound to all $d = O(n^{1/2}).$  A matching upper bound
of $2^{O(n)}$ on weight for $d=1$ has also long been known 
\cite{MyhillKautz:61}.
Our Theorem \ref{thm:mainptf} gives an upper bound 
which matches Beigel's lower bound (up to
logarithmic factors) for all $d = O(n^{1/3})$:
\begin{observation}
For any $d = O(n^{1/3})$ there is a polynomial threshold function of
degree $d$ and weight $2^{\tilde{O}(n/d^{2})}$ 
which computes $\mathrm{ODDMAXBIT}_{n}$. 
\end{observation}
\begin{proof}
Set $d = h^{1/2} \log h$ in Theorem~\ref{thm:mainptf}.  
The weight bound given by Theorem~\ref{thm:mainptf} 
is $2^{O({\frac {n \log^2 d}{d^2}} + d \log d)}$
which is $\tilde{O}(n/d^2)$ for $d = O(n^{1/3}).$
\end{proof} 

\medskip

Note that since the 
$\mathrm{ODDMAXBIT}_{n}$ function has a polynomial size DNF
(see Appendix \ref{ap:alt}), Beigel's lower bound gives a polynomial 
size DNF $f$ such that any degree $\tilde{O}(n^{1/3})$ polynomial
threshold function for $f$ must have weight
$2^{\tilde{\Omega}(n^{1/3})}$.
This suggests that the Expanded-Winnow algorithm cannot learn polynomial size
DNF in $2^{\tilde{O}(n^{1/3})}$ time from
$2^{n^{1/3 - \eps}}$ examples for any
$\eps > 0,$ and thus suggests that improving the sample complexity
of the DNF learning algorithm from \cite{KlivansServedio:01} while
maintaining its $2^{\tilde{O}(n^{1/3})}$ running time may be difficult.

\section{Learning Parity Functions} \label{sec:parity}

We first briefly review the standard
algorithm for learning parity functions.

The standard algorithm for learning parity functions works by viewing a
set of $m$ labelled examples as a set of $m$ linear equations over GF(2).
Each labelled example $(x,b)$ induces the equation
$\sum_{i: x_i = 1} a_{i} = b \bmod 2.$
Since the examples are labelled according to some parity function,
this parity function will be a consistent solution to the 
system of equations.  
Using Gaussian elimination it is possible to efficiently find a 
solution to the linear system, 
which yields a parity function consistent with all $m$ examples.
The following standard fact from learning theory
(often referred to as ``Occam's Razor'') shows that finding
a consistent hypothesis suffices to establish PAC learnability:

\begin{fact} \label{fact:OC}
Let $C$ be a concept class and $H$ a finite set of hypotheses. Set $m
= 1/\epsilon(\log |H| + \log 1/\delta)$ where $\epsilon$ and $\delta$
are the usual accuracy and confidence parameters for PAC learning.
Suppose that there
is an algorithm $A$ running in time $t$ which takes as input $m$
examples which are labelled according to some element of $C$ and outputs a 
hypothesis $h \in H$ consistent with these examples.  
Then $A$ is a PAC learning algorithm for $C$ with running time $t$
and sample complexity $m.$
\end{fact}
Consider using the above algorithm to learn an unknown 
parity of length at most $k.$
Even though there is a solution of weight at most $k$,
Gaussian elimination (applied to a system of $m$ equations in $n$
variables over GF(2)) may yield a solution of weight
as large as $\min(m,n).$  
Using Fact \ref{fact:OC} we thus obtain a sample complexity bound of 
$O(n)$ examples for learning a parity of length at most $k.$

We now present
a simple polynomial-time algorithm for learning an unknown parity
function on $k$ variables using $O(n^{1-1/k})$ examples.
To the best of our knowledge this is the first improvement on the
standard algorithm and analysis given above.

\begin{theorem} \label{thm:mainparity}

The class of all parity functions on at most $k$ variables is
learnable in polynomial time using $O(n^{1-1/k} \log n)$
examples. The hypothesis output by the learning algorithm
is a parity function on $O(n^{1-1/k}\log n)$ variables.

\end{theorem}

\begin{proof}
If $k = \Omega(\log n)$ then the standard algorithm suffices to 
prove the claimed bound.  We thus assume that $k = o(\log n)$.  

Let $H$ be the set of all parity functions of size at most $n^{1 - 1/k}$.
Note that $|H| \leq n^{n^{1 - 1/k}}$ so
$\log|H| \leq n^{1 - 1/k} \log n.$
Consider the following
algorithm:

\begin{enumerate}

\item Choose $m = 1/\epsilon (\log |H| + \log (1/\delta))$
examples. Express each example as a linear equation over $n$ variables
mod $2$ as described above.

\item Randomly choose a set of $n - n^{1-1/k}$ variables and assign
them the value $0$.

\item Use Gaussian elimination to attempt to solve the resulting system
of equations on the remaining $n^{1 - 1/k}$ variables.
If the system has a solution, output the corresponding parity
(of size at most $n^{1 - 1/k}$) as the hypothesis.
If the system has no solution, output ``FAIL.''

\end{enumerate}

If the simplified system of equations has a solution, 
then by Fact \ref{fact:OC} this solution is a good hypothesis.  
We will show that the simplified system has a solution with probability
$\Omega(1/n)$.  The theorem
follows  by repeating steps 2 and 3 of the above algorithm until
a solution is found (an expected $O(n)$ repetitions will suffice).

Let $V$ be the set of $k$ relevant variables on which the unknown
parity function depends. It is easy to see that as long as
no variable in $V$ is assigned a 0,
the resulting simplified system of equations will have a 
solution.  
Let $\ell = n^{1 - 1/k}.$
The probability that in Step 2 the $n - \ell$ variables chosen
do not include any variables in $V$ is exactly
${n - k \choose n - \ell} / {n \choose \ell}$
which equals
${n - k \choose \ell - k} / {n \choose \ell}.$  Expanding
binomial coefficients we have
\begin{equation} \label{eq:a}
{\frac {{n - k \choose \ell - k}}{{n \choose \ell}}} = 
\prod_{i=1}^{k} {\frac {\ell - k + i}{n -k + i}} 
> \left({\frac {\ell - k}{n - k}}\right)^k 
= 
\left({\frac \ell n}\right)^k 
\left({\frac {1 - {\frac k \ell}}{1 - {\frac k n}}}\right)^k
= 
{\frac 1 n} \cdot 
\left[\left(1 - {\frac k \ell}\right)\left(1 + {\frac {2k} n}\right)\right]^k.
\end{equation}
The bound $k = o(\log n)$ implies that 
$\left(1 - {\frac k \ell}\right)\left(1 + {\frac {2k} n}\right) > 
(1 - {\frac {3k} \ell}).$ Consequently 
(\ref{eq:a}) is at least
${\frac 1 n} \cdot \left(1 - {\frac {3k^2} {\ell}}\right) >
{\frac 1 {2n}}$ and the theorem is proved.
\end{proof}

\section{Future Work} \label{sec:future}

An obvious goal for future work is to improve our algorithmic results
for learning decision lists.  The question still remains:  can
decision lists of length $k$ be learned in poly$(n)$ time from
poly$(k,\log n)$ examples?  As a first step, one might attempt to
extend the tradeoffs we achieve:  is it possible to learn
decision lists of length $k$ in $n^{k^{1/2}}$ time from
poly$(k,\log n)$ examples?

Another goal is to extend our results for decision lists to broader
concept classes.  In particular, since decision lists are a special
case of linear threshold functions, it would be interesting to obtain analogues
of our algorithmic 
results for learning general linear threshold functions (independent of 
their weight).  We note here that
Goldmann {\em et al.} \cite{GHR:92} have given 
a linear threshold function over $\{-1,1\}^n$ for
which any polynomial threshold function must have weight
$2^{\Omega(n^{1/2})}$ regardless of its degree.  Moreover
Krause and Pudlak \cite{KrausePudlak:98} have shown that any Boolean
function which has a polynomial threshold function over $\{0,1\}^n$ of weight 
$w$ has a polynomial threshold function over $\{-1,1\}^n$ of weight
$n^2w^4.$  These results imply that {\em representational} results akin
to Theorem \ref{thm:ptf} for general linear threshold functions
must be quantitatively weaker than Theorem \ref{thm:ptf};
in particular, there is a linear threshold function over
$\{0,1\}^n$ with $k$ nonzero coefficients for which
{any} polynomial threshold function, regardless of degree, must have 
weight $2^{\Omega(k^{1/2})}.$

For parity functions, one challenge is to
learn parity functions on $k = \Theta(\log n)$ variables in polynomial time
using a sublinear number of examples.  Another challenge is to improve
the sample complexity of learning size $k$ parities from our
current bound of $O(n^{1 - 1/k}).$

\ignore{

Decision lists can be viewed as a special case of linear threshold
functions. For example, the alternating decision list (or
$\mathrm{ODDMAXBIT}_{n}$ function) is equal to the sign of $h =
\sum_{i=1}^{n} (-1)^{i} 2^{i}x_{i}$. The lower bound on the
$\mathrm{ODDMAXBIT}_{n}$ function due to Beigel shows that for an
arbitrary linear threshold function, we cannot construct polynomial
threshold functions of degree $d$ and weight $2^{o(n/d^{2})}.$

Here we observe that this lower bound on the weight and degree of
polynomial threshold functions computing general linear threshold
functions can be strengthened due to a result by Goldmann, Hastad, and
Razborov:

\begin{theorem} \cite{GHR:92}
There exists a linear threshold function $U$ defined on $4n^{2}$
variables such that if $U$ is written as a threshold of monomials then
the total weight of the threshold is $\Omega(2^{(n/2)} / \sqrt{n})$.
\end{theorem} 

\noindent The linear threshold function $U$ is the so-called Universal
Halfspace defined as follows:

\[ U_{n,m} = \sum_{i=1}^{n} \sum_{j=1}^{m} 2^{i}x_{ij}. \]

From this we conclude that to learn an arbitrary linear threshold
function on $n$ variables, V-Winnow will require
$\Omega(2^{\sqrt{n}})$ samples and time $\Omega(n^{\sqrt{n}})$. This
stands in contrast to the sample complexity and time complexity bounds
for learning decision lists.
}

\section{Acknowledgements} We thank Les Valiant for his observation
that Claim \ref{cla:outer} can be reinterpreted in terms of polynomial
threshold functions.  
We thank Jean Kwon for suggesting the Chebychev polynomial.

\bibliographystyle{plain} 
\bibliography{allrefs}

\appendix

\section{Alternate Proof of Corollary \ref{cor:outer}} \label{ap:alt}
The alternate proof of Corollary \ref{cor:outer} is based on the
observation that any decision list $L = 
(\ell_1,b_1),\dots,$ $(\ell_k,b_k),b_{k+1}$ of length $k$ has a
$k$-term DNF in which each term is a conjunction of at most
$k$ literals.  To see this, note that we obtain a DNF
for $L$ simply by taking the OR of all terms
$\overline{\ell}_1\overline{\ell}_2 \dots \overline{\ell}_{i-1}\ell_i$
for each $i$ such that $b_i = 1.$  Now we use the following result
from \cite{KlivansServedio:01}:
\begin{theorem} [Corollary 12 of \cite{KlivansServedio:01}]
Let $f$ be a DNF formula of $s$ terms, each of length at most $t.$
Then there is a polynomial threshold function for $f$ of degree
$O(\sqrt{t}\log s)$ and weight $t^{O(\sqrt{t}\log s)}.$
\end{theorem}
Applying this result to the DNF representation for $L,$ we immediately
obtain that there is a polynomial threshold function for $L$
which has degree $O(k^{1/2} \log k)$ and weight
$2^{O(k^{1/2} \log^2 k)}.$  (In Section \ref{subsec:inner}, though,
we need the construction given in our original proof of
Corollary \ref{cor:outer}.)

\end{document}